\documentclass[sigconf]{acmart}
\usepackage{mathtools}
\usepackage{subcaption}

\AtBeginDocument{%
  \providecommand\BibTeX{{%
    \normalfont B\kern-0.5em{\scshape i\kern-0.25em b}\kern-0.8em\TeX}}}

\setcopyright{acmcopyright}
\copyrightyear{2021}
\acmYear{2021}
\acmConference[VAM-HRI '21]{Virtual, Augmented and Mixed reality human-robot interaction Workshop}
\acmBooktitle{Virtual, Augmented and Mixed reality human-robot interaction Workshop}

\begin{document}

\title{HAIR: Head-mounted AR Intention Recognition}


\author{David Puljiz}
\affiliation{%
  \institution{Intelligent Process Automation and Robotics Lab (IPR), Institute for Anthropomatics and Robotics, Karlsruhe Institute of Technology}
  \city{Karlsruhe}
  \country{Germany}}
\email{david.puljiz@kit.edu}

\author{Bowen Zhou}
\affiliation{%
  \institution{Intelligent Process Automation and Robotics Lab (IPR), Institute for Anthropomatics and Robotics, Karlsruhe Institute of Technology}
  \city{Karlsruhe}
  \country{Germany}}

\author{Ke Ma}
\affiliation{%
  \institution{Intelligent Process Automation and Robotics Lab (IPR), Institute for Anthropomatics and Robotics, Karlsruhe Institute of Technology}
  \city{Karlsruhe}
  \country{Germany}}

\author{Bj\"orn Hein}
\affiliation{%
  \institution{Intelligent Process Automation and Robotics Lab (IPR), Institute for Anthropomatics and Robotics, Karlsruhe Institute of Technology}
  \city{Karlsruhe}
  \country{Germany}}

\renewcommand{\shortauthors}{Puljiz et al.}

\begin{abstract}
Human teams exhibit both implicit and explicit intention sharing. To further development of human-robot collaboration, intention recognition is crucial on both sides. Present approaches rely on a vast sensor suite on and around the robot to achieve intention recognition. This relegates intuitive human-robot collaboration purely to such bulky systems, which are inadequate for large-scale, real-world scenarios due to their complexity and cost. In this paper we propose an intention recognition system that is based purely on a portable head-mounted display. In addition robot intention visualisation is also supported. We present experiments to show the quality of our human goal estimation component and some basic interactions with an industrial robot. HAIR should raise the quality of interaction between robots and humans, instead of such interactions raising the hair on the necks of the human coworkers.       
\end{abstract}




\begin{teaserfigure}
  \includegraphics[width=\textwidth]{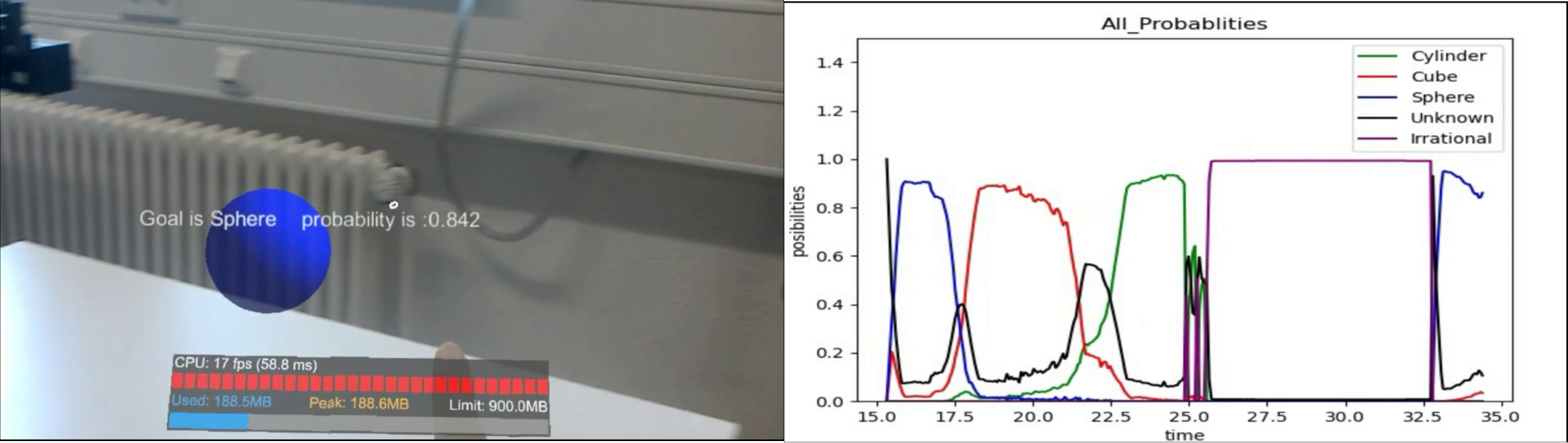}
  \caption{The view from the HoloLens on the left, showing the user looking at the goal "Sphere" and approaching it with the hand. On the right the current and previous probabilities. The user rotated from the first goal, the sphere, through the other two goals and back towards the sphere. To note is how the output proclaims the user irrational when they are facing away from all the defined goals}
  \label{fig:teaser}
\end{teaserfigure}

\maketitle

\section{Introduction}

\begin{figure*}[t!]
\centering
\vspace {5pt}
\subcaptionbox{}
[0.48\textwidth]{\includegraphics[width=0.48\textwidth]{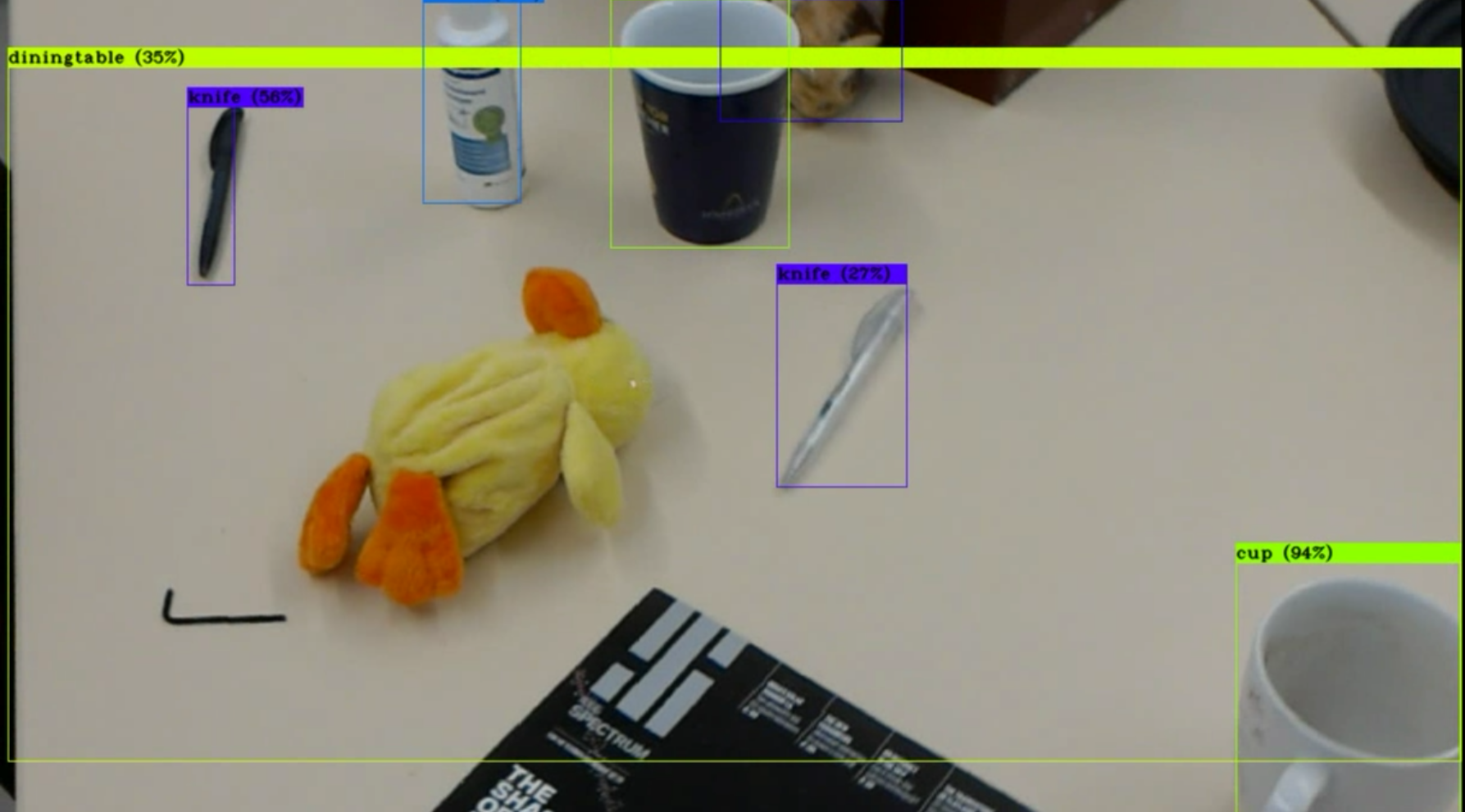}}
\subcaptionbox{}
[0.48\textwidth]{\includegraphics[width=0.48\textwidth]{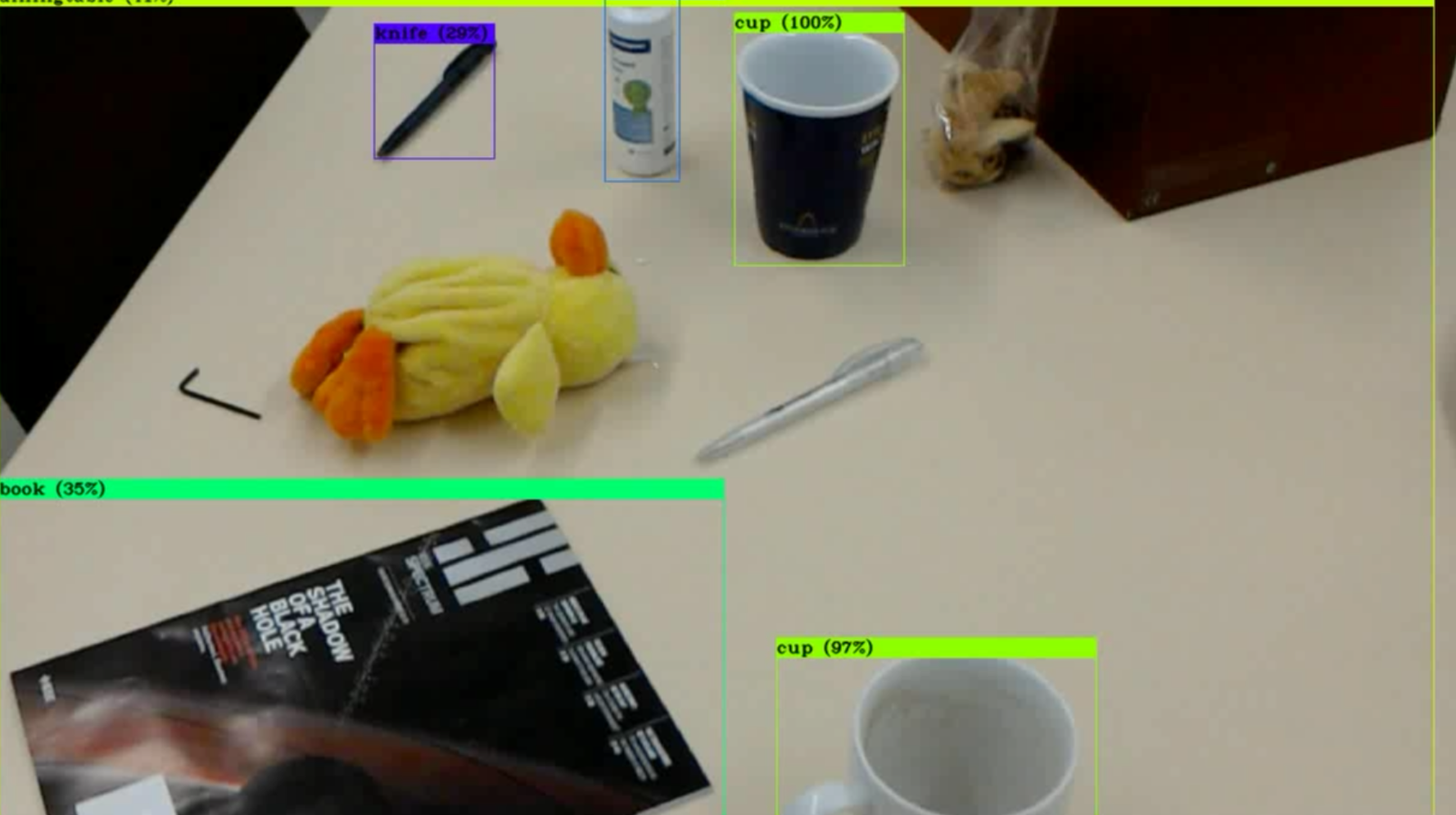}}
\caption[]{YOLOv4 classification of HoloLens camera data. Bounding box classification approaches together with the known HoloLens egomotion and depth data can be used to define the interaction objects/goals as well as to outsource part of the environmental sensing from the robot to the human.}
\label{fig:yolo}
\end{figure*}

Communicating intentions between members of a team is paramount for successful cooperation and task completion. Previous work in the field of Augmented reality (AR) human-robot interaction (HRI) focused on either improving robot programming \cite{quintero2018ar} or visualising robot motions \cite{walker2018motionar}. Although quite important for collaboration, such systems still lack the estimation of the human intention from the robot's side. Several such systems have been proposed, such as \cite{bascetta2011hir} where the human is tracked and their goal estimated inside a robot cell. Such systems however require a big overhead in complexity and cost of the robot cells. With the advent of Head-mounted Displays (HMDs) the possibility arises of a fully portable, completely worn system possessing both the robot intention visualisation and human intention estimation. \par       

A similar system based on a HMD and intended for human-robot collaborative task planning was presented by Chakraborti et al. \cite{Chakraborti2018task}. In the presented system, however, the human coworker had to explicitly select and reserve objects it wished to interact with, slowing down task execution and increasing physical and mental workloads on the human worker. \par

Here we propose a system that implicitly evaluates the intentions of the human, thus minimising the increase in workload. The proposed system is based on the Microsoft's HoloLens HMD and is aimed at a collaborative scenario between a single human and an industrial manipulator. The system is robot agnostic and completely portable requiring a very short set-up at the beginning of interaction. This guarantees that a single human worker can interface with multiple robots one after another, without the need for specialised robot cells or sensors around any of those robots. \par

The system takes as input the pose of the HMD in the world coordinate system, the position of the hand joints on the world coordinate system as well as a set of possible spatial goals, which can be added and removed during the interaction itself. The output is a set of probabilities of the goal the human wants to approach as well as the action they wish to perform. \par

This paper will present our current work and tests aimed mostly at having a robust goal estimation. To the best of the authors knowledge such an intention estimation algorithm using a completely worn system has not yet been developed. \par

\section{Methodology}

\subsection{Referencing}
First and foremost a common coordinate system must be establish between the HMD and the robotic manipulator. Referencing can be done in a variety of ways, perhaps the most popular is the use of QR codes or other preset visual markers \cite{krupke2018ar}. Although these offer continuous instead of one-shot referencing, as well as very good precision, they require a setup step that we would like to avoid. Manually selecting the robot base such as presented in \cite{quintero2018ar} is more flexible yet also more imprecise. We have proposed several referencing methods in \cite{puljiz2019referencing}, with the semi-automatic one, consisting of a rough user guess followed by a refinement step, offering the best balance between accuracy and computational time. The refinement step consists of filtering a point cloud captured with the HMD and using a registration algorithm to fit the model of the manipulator into the filtered point cloud, using the user guess as the start point of the registration algorithm. The user guess prevents the common problem of registration algorithms being stuck in the local minima, and we found that even a basic ICP algorithm performs a good job of refining the guess of the user. Another approach is a fully automatic one without a user guess. Such a referencing algorithm, similarly to our automatic method proposed in \cite{puljiz2019referencing}, was proposed by Ostantin et al. \cite{ostanin2020pointcloud}. It clusters the point cloud captured by the HMD using the DBSCAN clustering algorithm and then performs model matching between the clusters and a model of the robot. \par 

\subsection{Defining Spatial Goals}
Secondly the possible spatial goals of the human and the robot need to be defined. In case the robot does not possess the full map of its surroundings, the HMD can also provide that as we demonstrated in \cite{puljiz2020hololens}. This can also include possible goals and regions of interest such as the table or the conveyor belt. If goals are specific objects, here too the HMD can provide types and positions of those objects. One such possibility is through the use of bounding-box classifiers such as YOLOv4 \cite{bochkovskiy2020yolov4}. In Fig. \ref{fig:yolo}, one can see the result of running YOLOv4 on the HoloLens camera data. Having the egomotion data of the hololens, as well as data from the HMD's depth sensor, allows a full spatial definition of the objects and therefore possible goals. \par

The user should also be able to add and remove goals manually during the interaction step. Therefore the goal estimation algorithm was selected to allow such a modality. \par

\begin{figure}[ht]
\centering
\includegraphics[width=0.45\textwidth]{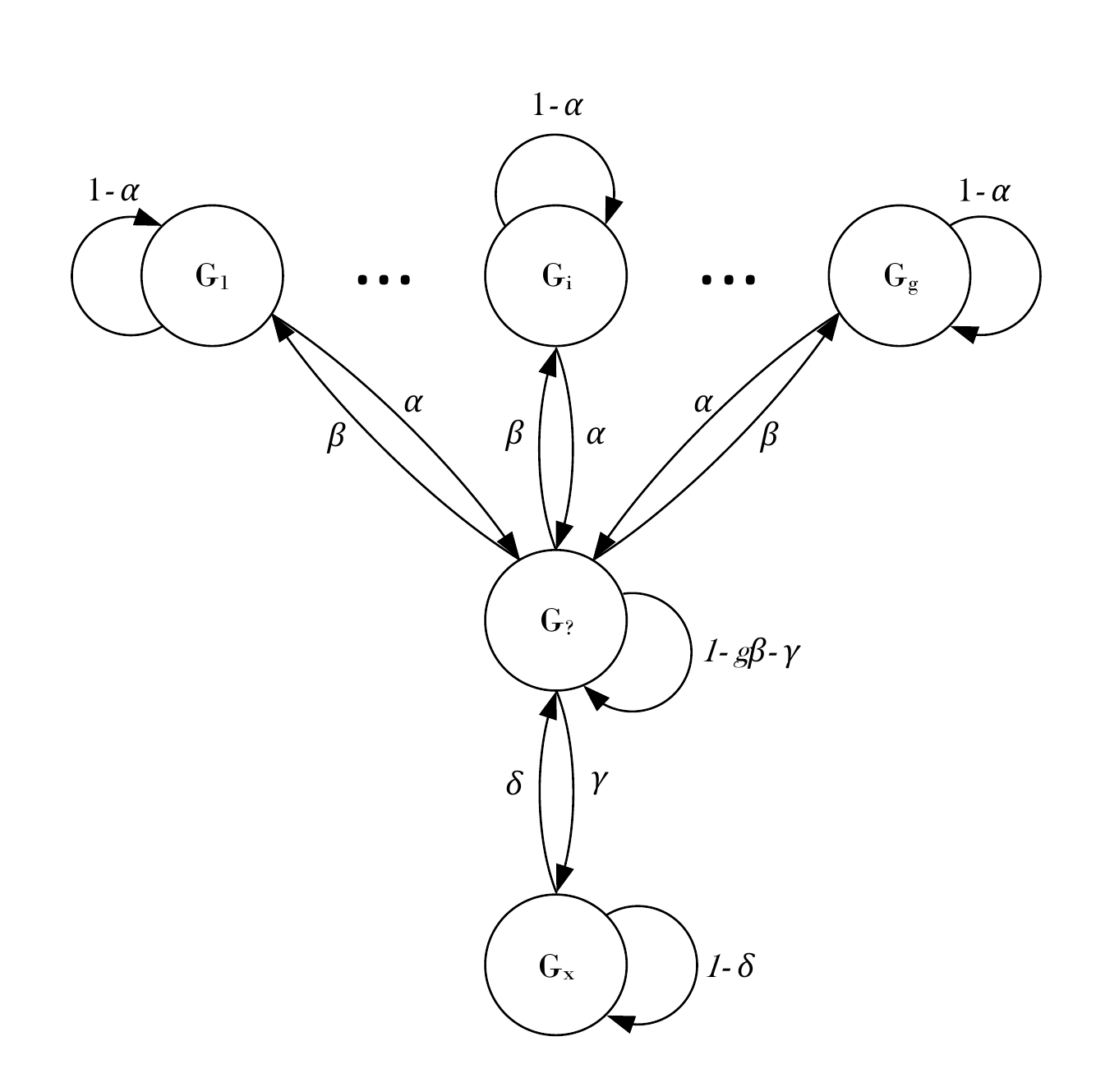}
\caption[]{The HMM states of the human goal intention estimation system as presented in \cite{PETKOVIC2019hir}. It consists of g goal states, a state of unknown intention $\textbf{G}_{?}$ and a state of the human acting irrationally $\textbf{G}_x$.}
    \label{fig:hmm}
\end{figure}

\subsection{Goal Estimation}
Finally, having a common coordinate system, mapped working environment and possible goals, one can infer the goals using a human intention recognition algorithm. We base our HIR algorithm on previous work by Petkovic et al. \cite{PETKOVIC2019hir} where a hidden Markov model framework was used to estimate the goal of the human in an automated warehouse. The approach is quite general and with minimal modification can be adapted to be used in our use case. In this section we will present a brief overview of the calculation, for more details please refer to the original paper. \par

Instead of the position of the human coworker as in the original paper, we consider the position of the hand in relation to goal objects. To simplify the calculations, we assume there is an almost straight line between the hand position and each goal object. By doing that we can forego the complex path planning step to determine the modulated distance and instead use the euclidean distance to calculate the vector \textbf{d} that represents the distance of the hand to each goal. As in \cite{PETKOVIC2019hir}, we define additional 32 points $p_{i}$ on a circle around the previous hand position $l'$ and a radius $r$ equal to the distance between the current $l$ and the previous $l'$ hand positions. We calculate the vector \textbf{d} for each point $p_{i}$ and append them to the modulated distance matrix \textbf{D}. \par

Additionally we consider the gaze validation \textbf{s} of the HMD. The motivation being that the user is more likely to look approximately towards the goal of the hand motion than towards other goals. The gaze validation is calculated as: \par

\begin{equation}
\textbf{s}_i =
\begin{dcases}
\textbf{g} \cdot \frac{\textbf{o}_i - \textbf{h}}{||\textbf{o}_i - \textbf{h}||}, & \textbf{g} \cdot \frac{\textbf{o}_i - \textbf{h}}{||\textbf{o}_i - \textbf{h}||} \geq 0 \\
0, & \text{otherwise.}
\end{dcases}
\label{eq:gaze}
\end{equation}

Where \textbf{g} is the HMD orientation in the world coordinate system, $\textbf{o}_i$ is the position of object i and \textbf{h} is the position of the HMD. We expand the motion validation vector \textbf{v} as follows: \par
\begin{equation}
\textbf{v} = \frac{\underset{1\leq i \leq n}\max{\textbf{D}_{ij}}-\textbf{d}}{\underset{1\leq i \leq n}\max{\textbf{D}_{ij}}-\underset{1\leq i \leq n}\min{\textbf{D}_{ij}}} \cdot \textbf{s}
\label{eq:validation}
\end{equation}

The rest follows exactly the algorithm described in \cite{PETKOVIC2019hir}. We use the same transition matrix with g goals $\textbf{T}^{g+2 \times g+2}$ defined as:

\begingroup
\renewcommand*{\arraycolsep}{5pt}
\begin{equation}
\textbf{T}=\begin{bmatrix}
1-\alpha& 0 & \dots & \alpha & 0 \\
0 & 1-\alpha &\dots & \alpha & 0 \\
\vdots & & \ddots & & \vdots \\
\beta&\beta& \dots & 1-g\beta-\gamma & \gamma \\
0 & 0 & \dots & \delta & 1-\delta \\
\end{bmatrix},
\label{eq:transition}
\end{equation}
\endgroup

This transition matrix corresponds to the hidden Markov model (HMM) architecture visible in Fig. \ref{fig:hmm}. The parameter $\alpha$ captures the worker tendency to change their mind, while the parameter couple $\beta$ and $\gamma$ set the threshold for estimating intention for each goal location. Increasing $\beta$ leads to quicker inference of worker's intentions and increasing $\gamma$ speeds up the decision making process. Parameter $\delta$ captures model's reluctance to return to estimating the other goal probabilities once it estimated that the worker is irrational. We performed several tests to determine the optimal values of these parameters which will be described in the "Experiments" section. \par

The  worker intention is estimated using the Viterbi algorithm \cite{viterbi}, which takes as inputs the hidden states set $S=\{G_1, ... G_g, G_?, G_x\} $, hidden state transition matrix $\textbf{T}$, initial state $\Pi$, sequence of observations $\textbf{O}$, and the emission matrix $\textbf{B}$. \par

The emission matrix $\textbf{B}$ is calculated using the motion validation vector \textbf{v}. Since the observation is the validation vector $\textbf{v}$ with continuous element values, the input to the Viterbi algorithm was modified by introducing an expandable emission matrix $\textbf{B}^{k \times g}$, where $k$ is the recorded number of observations, are functions of the observation value. Once a new validation vector $v$ is calculated, the emission matrix is expanded with the row $\textbf{B}'$, where the element $\textbf{B}'_{i}$ stores the probability of observing \textbf{v} from hidden state $G_i$.
The average of the last $m$ vectors \textbf{v} is also calculated and the maximum average value $\phi$ is selected. It is used as an indicator if the worker is behaving irrationally, i.e., is not moving towards any predefined goal. The value of the hyperparameter $m$ indicates how much evidence is to be collect before the worker is declared irrational. If the worker has been moving towards at least one goal in the last $m$ iterations ($\phi>0.5$),  $\textbf{B}'$ is calculated as:
\begin{equation}
B'=\zeta \cdot
\begin{bmatrix} \tanh(\textbf{v}) & \tanh(1-\Delta) & 0 \end{bmatrix}
\label{eq:Bmatrix},
\end{equation}
and otherwise as:
\begin{equation}
B'=\zeta \cdot
\begin{bmatrix} \boldsymbol{0}_{1\times g} & \tanh(0.1) & \tanh(1-\phi) \end{bmatrix},
\label{eq:Bmatrix2}
\end{equation}
where $\zeta$ is a normalising constant and $\Delta$ is calculated as the difference of the largest and second largest element of \textbf{v}. \par

Finally, the initial probabilities of worker's intentions are set as:
\begin{equation}
\Pi= \begin{bmatrix}
0 & \dots & 0 & 1 & 0
\end{bmatrix},
\label{eq:initial_desires}
\end{equation}
indicating that the initial state is $G_?$ and the model does not know which goal the worker desires the most.
The Viterbi algorithm outputs the most probable hidden state sequence and the probabilities $P(G_i)$ of each hidden state in each step. These probabilities are the worker's intention estimates.  

Goals can be added and removed during runtime as well making such a intention estimation framework quite flexible. \par

\begin{figure}[]
\centering
\includegraphics[width=0.45\textwidth]{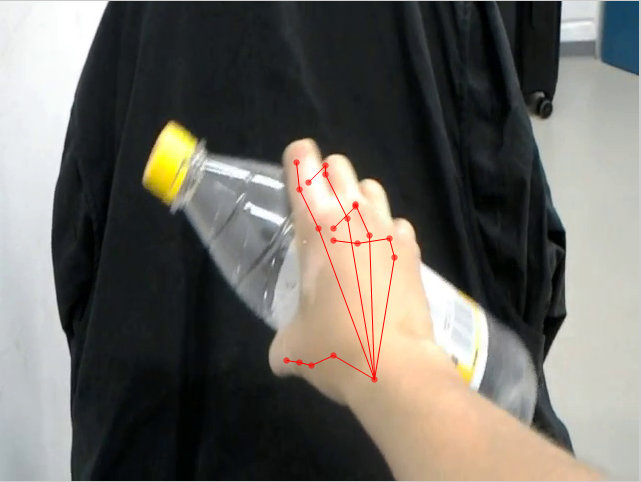}
\caption[]{The result of hand detection presented in \cite{GANeratedHands_CVPR2018} on HoloLens RGB camera data. One can see robust performance even during object handling.}
    \label{fig:hir}
\end{figure}

\subsection{Action Estimation}

Though estimating the goal of the human motion is extremely important for replanning robot motions to keep the interaction both safe and efficient, estimating actions the human wishes to perform could also bring additional information and flexibility to intention estimation systems. \par

Although the first generation of the HoloLens possesses inbuilt hand-tracking capabilities, these are quite limited and only four gestures can be tracked and classified. For a more robust hand following and classification we expanded the hand tracking by using the work presented in \cite{GANeratedHands_CVPR2018} on the HoloLens' RGB camera data. The algorithm tracks 21 hand joints and works with occlusions, surface contacts and object handling. \par

The detected hand joints are to classify actions - intention to grasp an object, grasped an object, pointing and stop. More actions can be classified in the future. The stop and pointing gestures are used as simple cues to control the robot. In addition, common gestures of fear or distress shall be classified as stop gestures, allowing the system indirect reaction to stress. \par

We presently only detect and use the right hand, however with a slight overhead the algorithm can detect both provided there is no significant overlap between them. \par

\begin{figure}[b]
\centering
\includegraphics[width=0.45\textwidth]{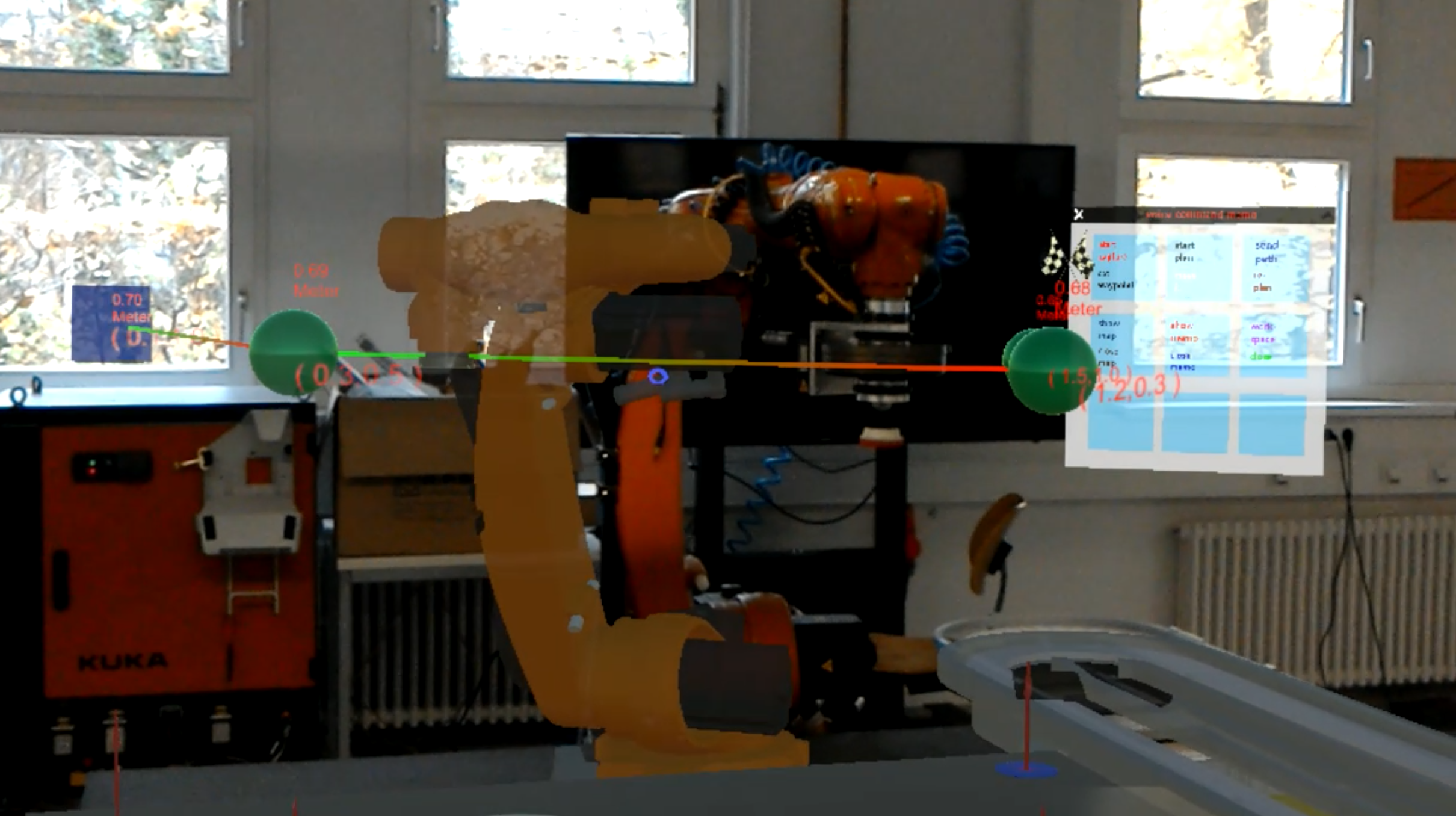}
\caption[]{An example of visualisation of the intentions of the robot to the human coworker - virtual execution with a holographic robot and plan visualisation.}
    \label{fig:virtual}
\end{figure}

\subsection{Robot Intention visualisation}
The benefits of HMDs extend also to visualising robot intention. Instead of adapting robot motions to make them more legible \cite{dragan2015legible}, one can use holograms to signal the desired goal. In \cite{williams2019ARcues} it was shown that holographic information is adequate to show the goal of the robot, and even solve ambiguities if intention is expressed via synthesised voice. General motion intent can also be effectively visualised using holographic cues \cite{walker2018motionar}. In our work we chose to indicate the goal via a hologram containing a 3D sound source (spatial sound), as well as virtual execution - having a hologram of the robot execute the motion before the real robot performs it, such as shown in Fig. \ref{fig:virtual}.

\begin{figure*}[t!]
\centering
\includegraphics[width=0.95\textwidth]{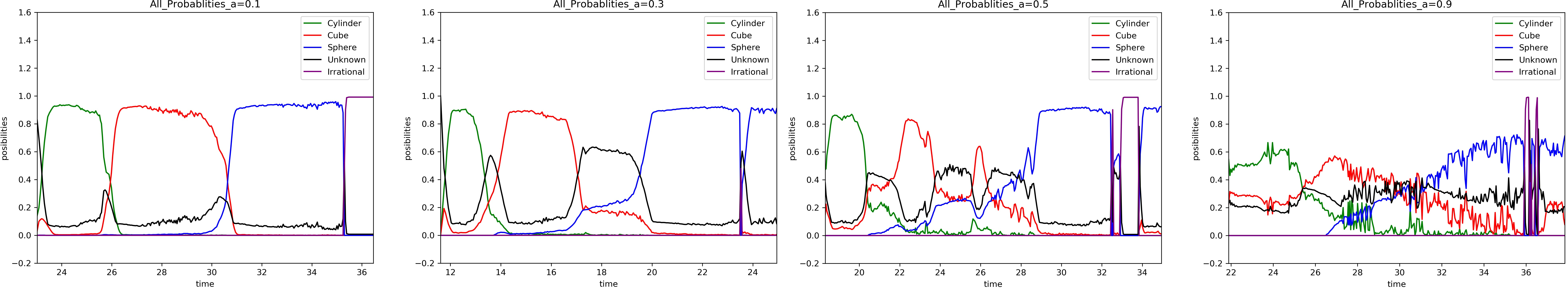}
\caption[]{The effect of increasing the parameter $\alpha$. The parameter captures the worker tendency to change their mind. A low $\alpha$ will make the algorithm "too sure" about the intention, while a too high $\alpha$ produces a chaotic and unusable output. }
\label{fig:alpha}
\end{figure*}

\section{Experiments}
The experiments were aimed at testing the performance of the goal intention estimation. We used three goals in a circular pattern, from left to right - a green cylinder, a red cube and a blue sphere, as shown in Fig. \ref{fig:hir}. \par 

The first set of experiments was aimed to find the optimal set of parameters $\alpha$, $\beta$, $\gamma$ and $\delta$ for our use case. Here we looked at the goal states and the transitions between them. The path was a simple left to right one, first going towards the green cylinder, then the red cube and finally the blue sphere. Figure Fig. \ref{fig:alpha} shows the behaviour of the parameter $\alpha$. A low value makes the algorithm too certain and almost does not spend time in the unknown state, while a high value makes the estimated goals jump too much. The optimal value of the parameter $\alpha$ was therefore set to $\alpha=0.3$. \par

\begin{figure*}[t!]
\centering
\includegraphics[width=0.95\textwidth]{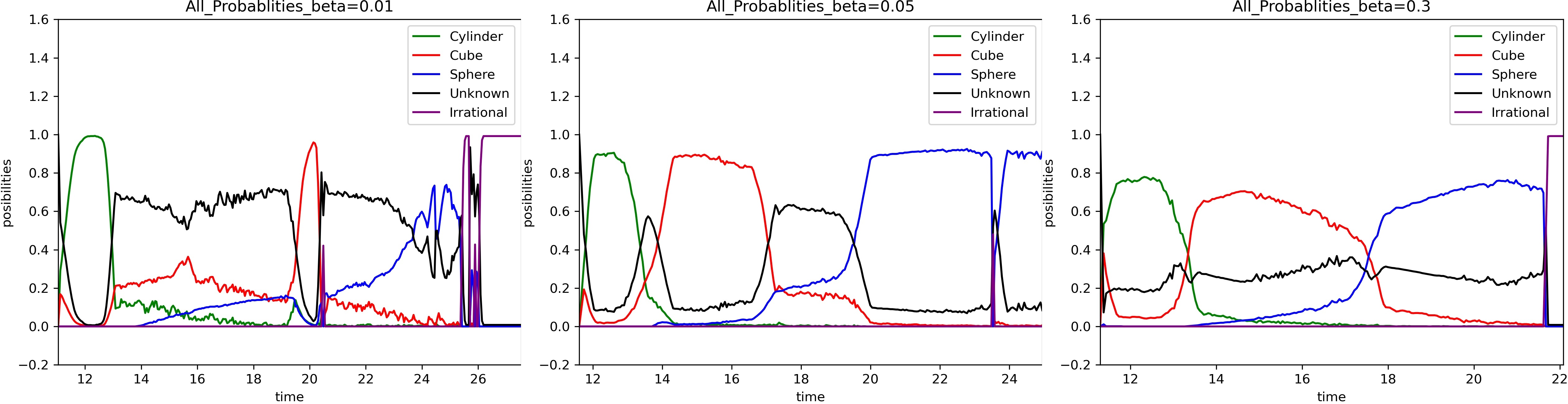}
\caption[]{The effect of increasing the parameter $\beta$. The parameter couple $\beta$ and $\gamma$ set the threshold for estimating intentions for each goal location. A low $\beta$ will make the algorithm estimate the unknown intention too much rendering it unusable , while a too high $\beta$ lowers the general certainty but eliminates the insecurities between transitions which is an unwanted behaviour.}
\label{fig:beta}
\end{figure*}

With the parameter $\alpha$ set, we tested the behaviour of changing the parameter $\beta$. A low $\beta$maintains the unknown intention state too long, while a high $\beta$ lowers the general certainty but eliminates the insecurities between state transitions which is an unwanted behaviour. The value of $\beta$ was set to $\beta=0.05$. In Fig. \ref{fig:beta} one can see the behaviour of changing the parameter $\beta$. \par

The parameters $\gamma$ and $\delta$ did not significantly influence the outputs and were kept at the same value as in \cite{PETKOVIC2019hir}, namely $\gamma=0.05$ and $\delta=0.1$. \par

\begin{figure*}[t!]
\centering
\includegraphics[width=0.3\textwidth]{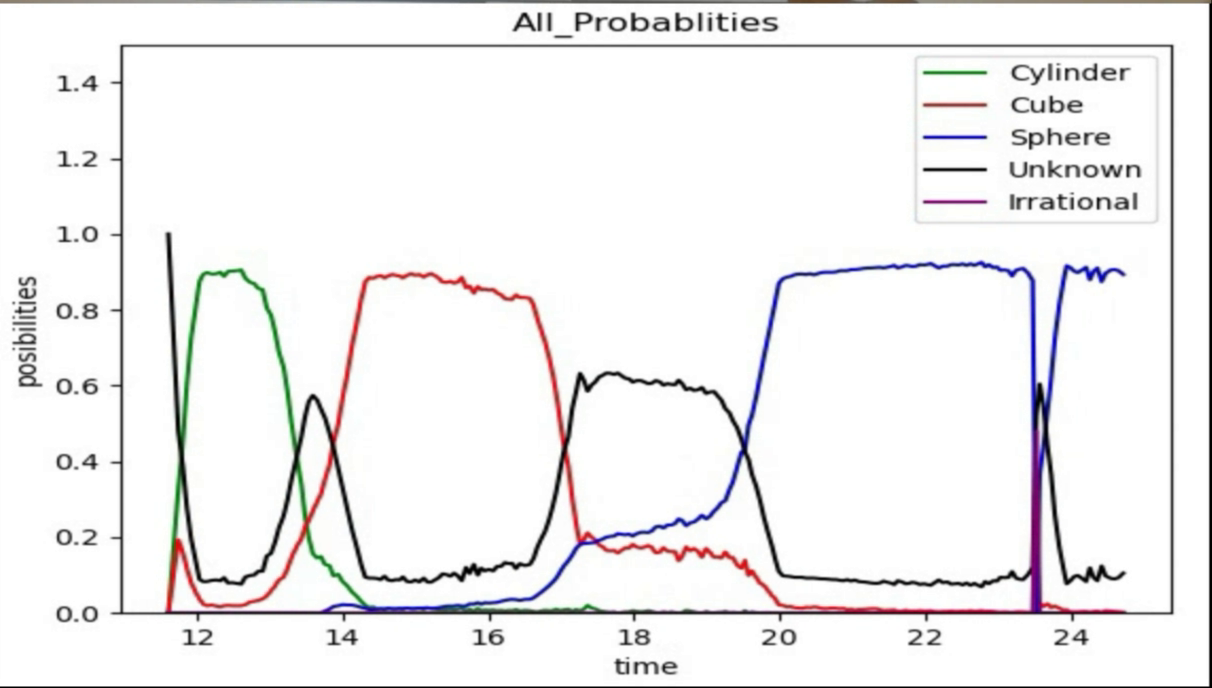}
\includegraphics[width=0.3\textwidth]{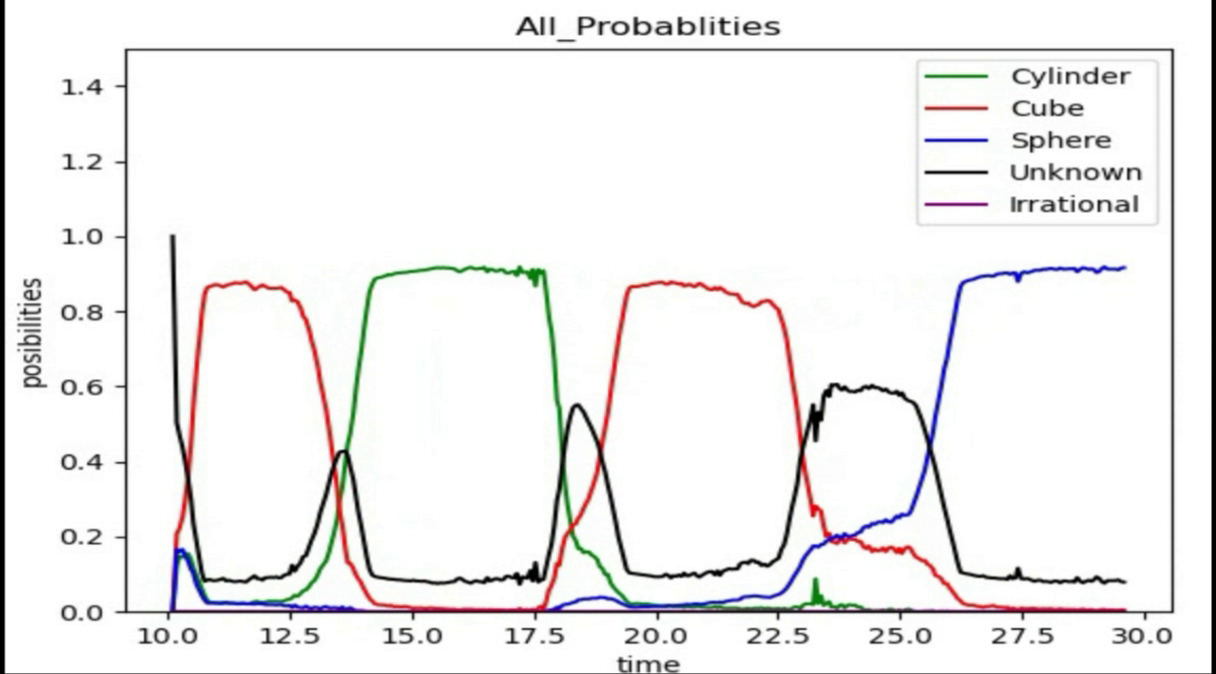}
\includegraphics[width=0.3\textwidth]{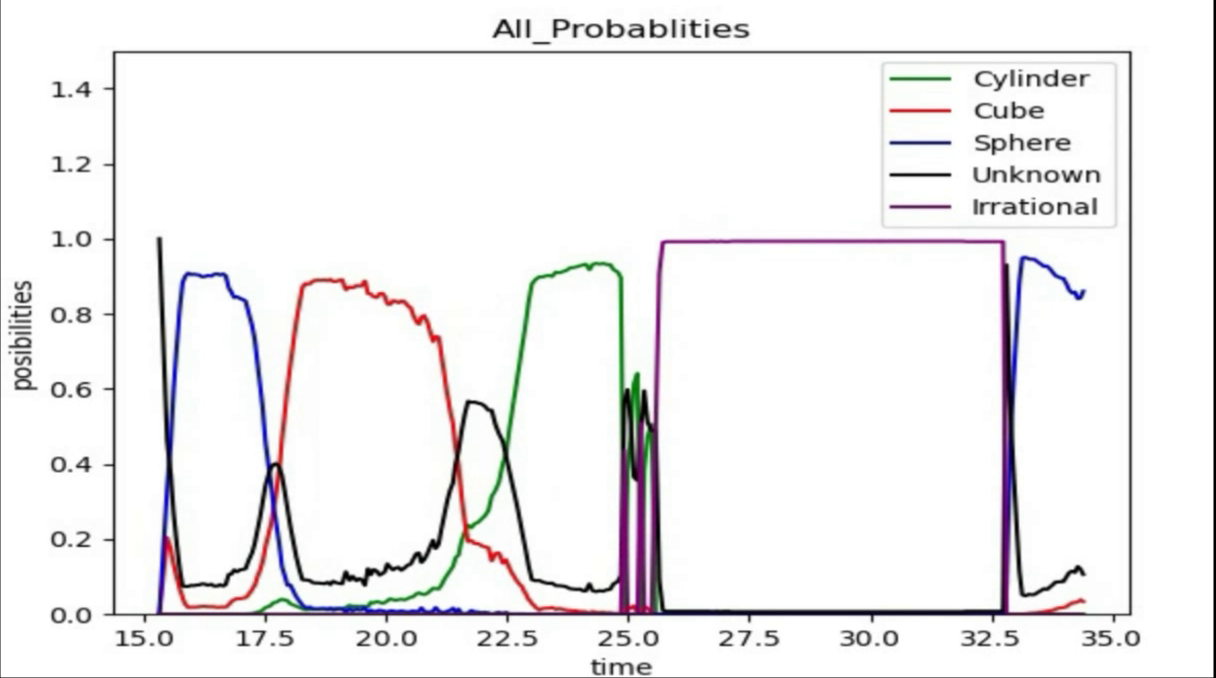}
\caption[]{Three tests with different goal order. On the left, the user was selecting goals left to right - cylinder, cube then sphere. In the middle the user starts with the cube, moves to the cylinder, back to the cube and finally goes to the sphere. In the test on the right the starting goal is sphere, then cube, then cylinder, after which the user turns around completely and ends back on the sphere.}
\label{fig:tests}
\end{figure*}

With the parameters set we examined how the algorithm behaves with different sequences of goals. The results are visible in Fig. \ref{fig:tests}. \par

The first test on the left is is a simple sequence of goals from left to right. One can notice the algorithm goes into the state of unknown intention during the transitions. The slower the transition the longer the unknown state. One can also see the small drop near the end when the hand tracking was lost and the gaze was not directly towards the sphere. \par

The second experiment starts with the middle goal, the cube, then goes left to the cylinder, back to the cube and then towards the sphere. One can see that the transition from cylinder to cube lasts slightly longer than from cube to cylinder. This is due to the fact that the algorithm is reluctant to estimate an already visited or skipped goal. One can also see the long transition between the cube and the sphere, as the algorithm prefers the goal that has already been visited two times. This shows that the estimation follows our intuition. \par

Finally, the third experiment shows what happens when the user does a complete rotation and faces away from all of the three goals. Again the algorithm performs quite intuitively and proclaims the user "irrational" as all the possible goals were completely on the other side. \par 

Additionally, we tested simple interactions between an industrial manipulator and the human user. In the first one the robot was selecting goals and randomly. Should the goal intention estimation detect that the human is moving to the same goal the robot would stop and select a new goal. Additionally we used the same framework to navigate the manipulator to the estimated goal of the human, proving that the framework can also be beneficial in teleportation scenarios. \par

\begin{figure}[b]
\centering
\includegraphics[width=0.45\textwidth]{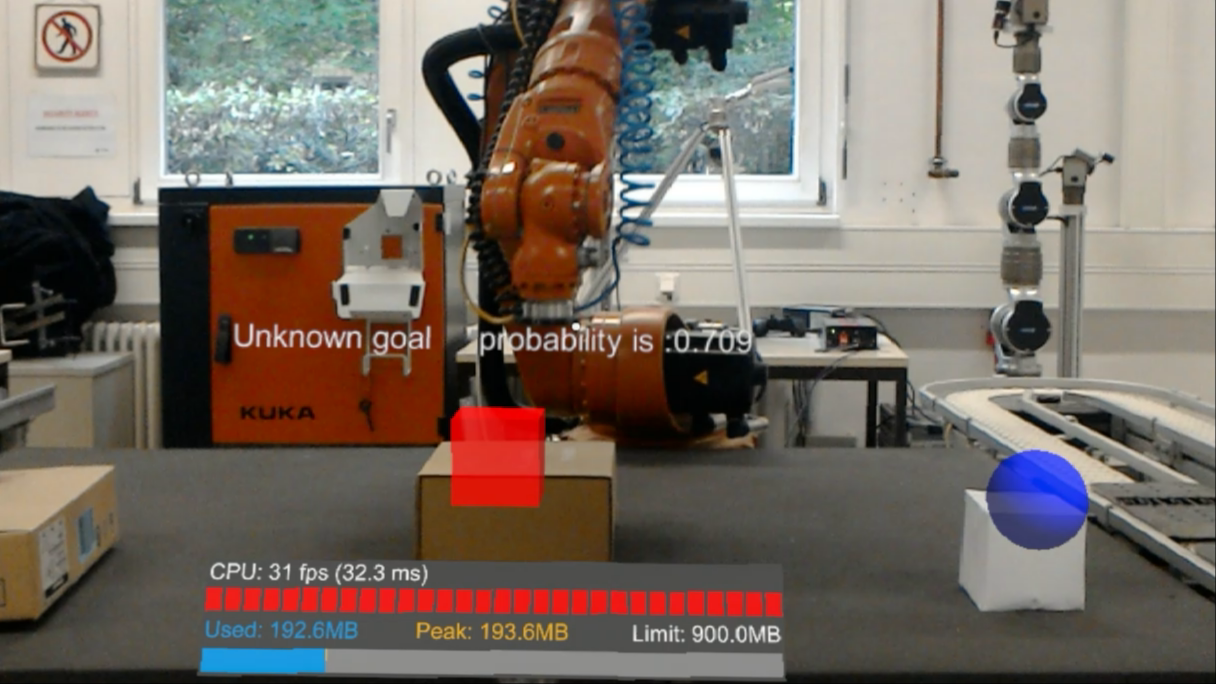}
\caption[]{The interaction setup with an industrial robot, the three spatial goals are represented by colour and shape. In this experiment the user took direct control of the robot and the human intention estimation is used to detect  to which object does the user wish the robot to move, illustrating another use case of intention estimation.}
    \label{fig:hir}
\end{figure}

\section{Conclusion and Impact}

In this work we presented a completely portable, robot agnostic system for intention estimation and visualisation for human-robot collaboration scenarios. Our system does not require any special set-up or sensors on or around the robot and is capable of both estimating the human coworkers goals and actions as well as visualise the goals and intentions of the robot coworker. \par

Having an intention estimation system, in addition to explicit intention declarations, can greatly reduce the mental and physical workload on the user, while providing constant, information rich data to the robot, thereby improving the safety and efficiency of robot motions. \par 
We have shown that the goal prediction part of the HAIR system works as intended and indeed the goal intentions estimated follow a reasoning that humans might find intuitive and agree with. \par 

Predicting the goal and motion of the human coworker can increase both safety for the human and the efficiency of robot motions. The goal estimation system \cite{PETKOVIC2019hir} was integrated into a mobile robot fleet management system of a simulated automated warehouse. In \cite{petkovic2020human} it was shown that the proposed system markedly improved warehouse efficiency compared to no goal estimation or even a simplistic one. It is to be expected that such an efficiency increase would also be observed in interactions with a robotic manipulator. Further testing is going to be needed however, to support that claim. \par 

Likewise the action estimation component as well as the entire system needs to be evaluated in user studies. More specifically the change in mental and physical workloads between various intention sharing modalities is of great interest and quite important in proving the claims that the intention estimation algorithms presented here significantly decrease the workload compared to explicitly stating the goals. \par 

As HMDs become ever more common, and the amount of robot coworkers per human coworker continues to increase, having intuitive HRI using systems that are cheap, simple and portable becomes essential. Lowering the complexity and price of each robot by exploiting wearables will lead to a wider use of robots and increased human-robot collaboration. We hope that the research presented here provides the first stepping stones towards such a system. HAIR should raise the quality of interaction between robots and humans, instead of such interactions raising the hair on the necks of the human coworkers. \par  

% \begin{acks}
% This work has been supported by... 
% \end{acks}

\bibliographystyle{ACM-Reference-Format}
\bibliography{sample-base}


\begin{thebibliography}{15}

%%% the Web address.

\ifx \showCODEN    \undefined \def \showCODEN     #1{\unskip}     \fi
\ifx \showDOI      \undefined \def \showDOI       #1{#1}\fi
\ifx \showISBNx    \undefined \def \showISBNx     #1{\unskip}     \fi
\ifx \showISBNxiii \undefined \def \showISBNxiii  #1{\unskip}     \fi
\ifx \showISSN     \undefined \def \showISSN      #1{\unskip}     \fi
\ifx \showLCCN     \undefined \def \showLCCN      #1{\unskip}     \fi
\ifx \shownote     \undefined \def \shownote      #1{#1}          \fi
\ifx \showarticletitle \undefined \def \showarticletitle #1{#1}   \fi
\ifx \showURL      \undefined \def \showURL       {\relax}        \fi
% invisible to TeX
\providecommand\bibfield[2]{#2}
\providecommand\bibinfo[2]{#2}
\providecommand\natexlab[1]{#1}
\providecommand\showeprint[2][]{arXiv:#2}

\bibitem[\protect\citeauthoryear{{Bascetta}, {Ferretti}, {Rocco}, {Ardö},
  {Bruyninckx}, {Demeester}, and {Di Lello}}{{Bascetta} et~al\mbox{.}}{2011}]%
        {bascetta2011hir}
\bibfield{author}{\bibinfo{person}{L. {Bascetta}}, \bibinfo{person}{G.
  {Ferretti}}, \bibinfo{person}{P. {Rocco}}, \bibinfo{person}{H. {Ardö}},
  \bibinfo{person}{H. {Bruyninckx}}, \bibinfo{person}{E. {Demeester}}, {and}
  \bibinfo{person}{E. {Di Lello}}.} \bibinfo{year}{2011}\natexlab{}.
\newblock \showarticletitle{Towards safe human-robot interaction in robotic
  cells: An approach based on visual tracking and intention estimation}. In
  \bibinfo{booktitle}{\emph{2011 IEEE/RSJ International Conference on
  Intelligent Robots and Systems}}. \bibinfo{pages}{2971--2978}.
\newblock
\urldef\tempurl%
\url{https://doi.org/10.1109/IROS.2011.6094642}
\showDOI{\tempurl}


\bibitem[\protect\citeauthoryear{Bochkovskiy, Wang, and Liao}{Bochkovskiy
  et~al\mbox{.}}{2020}]%
        {bochkovskiy2020yolov4}
\bibfield{author}{\bibinfo{person}{Alexey Bochkovskiy},
  \bibinfo{person}{Chien-Yao Wang}, {and} \bibinfo{person}{Hong-Yuan~Mark
  Liao}.} \bibinfo{year}{2020}\natexlab{}.
\newblock \showarticletitle{YOLOv4: Optimal Speed and Accuracy of Object
  Detection}.
\newblock \bibinfo{journal}{\emph{arXiv preprint arXiv:2004.10934}}
  (\bibinfo{year}{2020}).
\newblock


\bibitem[\protect\citeauthoryear{{Chakraborti}, {Sreedharan}, {Kulkarni}, and
  {Kambhampati}}{{Chakraborti} et~al\mbox{.}}{2018}]%
        {Chakraborti2018task}
\bibfield{author}{\bibinfo{person}{T. {Chakraborti}}, \bibinfo{person}{S.
  {Sreedharan}}, \bibinfo{person}{A. {Kulkarni}}, {and} \bibinfo{person}{S.
  {Kambhampati}}.} \bibinfo{year}{2018}\natexlab{}.
\newblock \showarticletitle{Projection-Aware Task Planning and Execution for
  Human-in-the-Loop Operation of Robots in a Mixed-Reality Workspace}. In
  \bibinfo{booktitle}{\emph{2018 IEEE/RSJ International Conference on
  Intelligent Robots and Systems (IROS)}}. \bibinfo{pages}{4476--4482}.
\newblock
\urldef\tempurl%
\url{https://doi.org/10.1109/IROS.2018.8593830}
\showDOI{\tempurl}


\bibitem[\protect\citeauthoryear{{Dragan}, {Bauman}, {Forlizzi}, and
  {Srinivasa}}{{Dragan} et~al\mbox{.}}{2015}]%
        {dragan2015legible}
\bibfield{author}{\bibinfo{person}{A.~D. {Dragan}}, \bibinfo{person}{S.
  {Bauman}}, \bibinfo{person}{J. {Forlizzi}}, {and} \bibinfo{person}{S.~S.
  {Srinivasa}}.} \bibinfo{year}{2015}\natexlab{}.
\newblock \showarticletitle{Effects of Robot Motion on Human-Robot
  Collaboration}. In \bibinfo{booktitle}{\emph{2015 10th ACM/IEEE International
  Conference on Human-Robot Interaction (HRI)}}. \bibinfo{pages}{51--58}.
\newblock


\bibitem[\protect\citeauthoryear{{Forney}}{{Forney}}{1973}]%
        {viterbi}
\bibfield{author}{\bibinfo{person}{G.~D. {Forney}}.}
  \bibinfo{year}{1973}\natexlab{}.
\newblock \showarticletitle{The viterbi algorithm}.
\newblock \bibinfo{journal}{\emph{Proc. IEEE}} \bibinfo{volume}{61},
  \bibinfo{number}{3} (\bibinfo{year}{1973}), \bibinfo{pages}{268--278}.
\newblock
\urldef\tempurl%
\url{https://doi.org/10.1109/PROC.1973.9030}
\showDOI{\tempurl}


\bibitem[\protect\citeauthoryear{{Krupke}, {Steinicke}, {Lubos}, {Jonetzko},
  {Görner}, and {Zhang}}{{Krupke} et~al\mbox{.}}{2018}]%
        {krupke2018ar}
\bibfield{author}{\bibinfo{person}{D. {Krupke}}, \bibinfo{person}{F.
  {Steinicke}}, \bibinfo{person}{P. {Lubos}}, \bibinfo{person}{Y. {Jonetzko}},
  \bibinfo{person}{M. {Görner}}, {and} \bibinfo{person}{J. {Zhang}}.}
  \bibinfo{year}{2018}\natexlab{}.
\newblock \showarticletitle{Comparison of Multimodal Heading and Pointing
  Gestures for Co-Located Mixed Reality Human-Robot Interaction}. In
  \bibinfo{booktitle}{\emph{2018 IEEE/RSJ International Conference on
  Intelligent Robots and Systems (IROS)}}. \bibinfo{pages}{1--9}.
\newblock
\urldef\tempurl%
\url{https://doi.org/10.1109/IROS.2018.8594043}
\showDOI{\tempurl}


\bibitem[\protect\citeauthoryear{Mueller, Bernard, Sotnychenko, Mehta, Sridhar,
  Casas, and Theobalt}{Mueller et~al\mbox{.}}{2018}]%
        {GANeratedHands_CVPR2018}
\bibfield{author}{\bibinfo{person}{Franziska Mueller}, \bibinfo{person}{Florian
  Bernard}, \bibinfo{person}{Oleksandr Sotnychenko}, \bibinfo{person}{Dushyant
  Mehta}, \bibinfo{person}{Srinath Sridhar}, \bibinfo{person}{Dan Casas}, {and}
  \bibinfo{person}{Christian Theobalt}.} \bibinfo{year}{2018}\natexlab{}.
\newblock \showarticletitle{GANerated Hands for Real-Time 3D Hand Tracking from
  Monocular RGB}. In \bibinfo{booktitle}{\emph{Proceedings of Computer Vision
  and Pattern Recognition ({CVPR})}}. \bibinfo{numpages}{11}~pages.
\newblock
\urldef\tempurl%
\url{https://handtracker.mpi-inf.mpg.de/projects/GANeratedHands/}
\showURL{%
\tempurl}


\bibitem[\protect\citeauthoryear{{Ostanin}, {Mikhel}, {Evlampiev},
  {Skvortsova}, and {Klimchik}}{{Ostanin} et~al\mbox{.}}{2020}]%
        {ostanin2020pointcloud}
\bibfield{author}{\bibinfo{person}{M. {Ostanin}}, \bibinfo{person}{S.
  {Mikhel}}, \bibinfo{person}{A. {Evlampiev}}, \bibinfo{person}{V.
  {Skvortsova}}, {and} \bibinfo{person}{A. {Klimchik}}.}
  \bibinfo{year}{2020}\natexlab{}.
\newblock \showarticletitle{Human-robot interaction for robotic manipulator
  programming in Mixed Reality}. In \bibinfo{booktitle}{\emph{2020 IEEE
  International Conference on Robotics and Automation (ICRA)}}.
  \bibinfo{pages}{2805--2811}.
\newblock
\urldef\tempurl%
\url{https://doi.org/10.1109/ICRA40945.2020.9196965}
\showDOI{\tempurl}


\bibitem[\protect\citeauthoryear{Petković, Hvězda, Rybecký, Marković,
  Kulich, Přeučil, and Petrović}{Petković et~al\mbox{.}}{2020}]%
        {petkovic2020human}
\bibfield{author}{\bibinfo{person}{Tomislav Petković}, \bibinfo{person}{Jakub
  Hvězda}, \bibinfo{person}{Tomáš Rybecký}, \bibinfo{person}{Ivan
  Marković}, \bibinfo{person}{Miroslav Kulich}, \bibinfo{person}{Libor
  Přeučil}, {and} \bibinfo{person}{Ivan Petrović}.}
  \bibinfo{year}{2020}\natexlab{}.
\newblock \bibinfo{title}{Human Intention Recognition for Human Aware Planning
  in Integrated Warehouse Systems}.
\newblock
\newblock
\showeprint[arxiv]{2005.11202}~[cs.RO]


\bibitem[\protect\citeauthoryear{Petković, Puljiz, Marković, and
  Hein}{Petković et~al\mbox{.}}{2019}]%
        {PETKOVIC2019hir}
\bibfield{author}{\bibinfo{person}{Tomislav Petković}, \bibinfo{person}{David
  Puljiz}, \bibinfo{person}{Ivan Marković}, {and} \bibinfo{person}{Björn
  Hein}.} \bibinfo{year}{2019}\natexlab{}.
\newblock \showarticletitle{Human intention estimation based on hidden Markov
  model motion validation for safe flexible robotized warehouses}.
\newblock \bibinfo{journal}{\emph{Robotics and Computer-Integrated
  Manufacturing}}  \bibinfo{volume}{57} (\bibinfo{year}{2019}),
  \bibinfo{pages}{182 -- 196}.
\newblock
\showISSN{0736-5845}
\urldef\tempurl%
\url{https://doi.org/10.1016/j.rcim.2018.11.004}
\showDOI{\tempurl}


\bibitem[\protect\citeauthoryear{Puljiz, Krebs, B{\"o}sing, and Hein}{Puljiz
  et~al\mbox{.}}{2020}]%
        {puljiz2020hololens}
\bibfield{author}{\bibinfo{person}{David Puljiz}, \bibinfo{person}{Franziska
  Krebs}, \bibinfo{person}{Fabian B{\"o}sing}, {and} \bibinfo{person}{Bj{\"o}rn
  Hein}.} \bibinfo{year}{2020}\natexlab{}.
\newblock \showarticletitle{What the HoloLens Maps Is Your Workspace: Fast
  Mapping and Set-up of Robot Cells via Head Mounted Displays and Augmented
  Reality}.
\newblock \bibinfo{journal}{\emph{arXiv preprint arXiv:2005.12651}}
  (\bibinfo{year}{2020}).
\newblock


\bibitem[\protect\citeauthoryear{Puljiz, Riesterer, Hein, and
  Kr{\"o}ger}{Puljiz et~al\mbox{.}}{2019}]%
        {puljiz2019referencing}
\bibfield{author}{\bibinfo{person}{David Puljiz}, \bibinfo{person}{Katharina~S
  Riesterer}, \bibinfo{person}{Bj{\"o}rn Hein}, {and} \bibinfo{person}{Torsten
  Kr{\"o}ger}.} \bibinfo{year}{2019}\natexlab{}.
\newblock \showarticletitle{Referencing between a Head-Mounted Device and
  Robotic Manipulators}. In \bibinfo{booktitle}{\emph{Proceedings of the 2nd
  Workshop on Virtual, Mixed and Augmented Reality Human.Robot Interaction, HRI
  2019}}.
\newblock
\urldef\tempurl%
\url{http://arxiv.org/abs/1904.02480}
\showURL{%
\tempurl}


\bibitem[\protect\citeauthoryear{{Quintero}, {Li}, {Pan}, {Chan}, {Machiel Van
  der Loos}, and {Croft}}{{Quintero} et~al\mbox{.}}{2018}]%
        {quintero2018ar}
\bibfield{author}{\bibinfo{person}{C.~P. {Quintero}}, \bibinfo{person}{S.
  {Li}}, \bibinfo{person}{M.~K. {Pan}}, \bibinfo{person}{W.~P. {Chan}},
  \bibinfo{person}{H.~F. {Machiel Van der Loos}}, {and} \bibinfo{person}{E.
  {Croft}}.} \bibinfo{year}{2018}\natexlab{}.
\newblock \showarticletitle{Robot Programming Through Augmented Trajectories in
  Augmented Reality}. In \bibinfo{booktitle}{\emph{2018 IEEE/RSJ International
  Conference on Intelligent Robots and Systems (IROS)}}.
  \bibinfo{pages}{1838--1844}.
\newblock
\urldef\tempurl%
\url{https://doi.org/10.1109/IROS.2018.8593700}
\showDOI{\tempurl}


\bibitem[\protect\citeauthoryear{Walker, Hedayati, Lee, and Szafir}{Walker
  et~al\mbox{.}}{2018}]%
        {walker2018motionar}
\bibfield{author}{\bibinfo{person}{Michael Walker}, \bibinfo{person}{Hooman
  Hedayati}, \bibinfo{person}{Jennifer Lee}, {and} \bibinfo{person}{Daniel
  Szafir}.} \bibinfo{year}{2018}\natexlab{}.
\newblock \showarticletitle{Communicating Robot Motion Intent with Augmented
  Reality}. In \bibinfo{booktitle}{\emph{Proceedings of the 2018 ACM/IEEE
  International Conference on Human-Robot Interaction}} (Chicago, IL, USA)
  \emph{(\bibinfo{series}{HRI '18})}. \bibinfo{publisher}{Association for
  Computing Machinery}, \bibinfo{address}{New York, NY, USA},
  \bibinfo{pages}{316–324}.
\newblock
\showISBNx{9781450349536}
\urldef\tempurl%
\url{https://doi.org/10.1145/3171221.3171253}
\showDOI{\tempurl}


\bibitem[\protect\citeauthoryear{{Williams}, {Bussing}, {Cabrol}, {Boyle}, and
  {Tran}}{{Williams} et~al\mbox{.}}{2019}]%
        {williams2019ARcues}
\bibfield{author}{\bibinfo{person}{T. {Williams}}, \bibinfo{person}{M.
  {Bussing}}, \bibinfo{person}{S. {Cabrol}}, \bibinfo{person}{E. {Boyle}},
  {and} \bibinfo{person}{N. {Tran}}.} \bibinfo{year}{2019}\natexlab{}.
\newblock \showarticletitle{Mixed Reality Deictic Gesture for Multi-Modal Robot
  Communication}. In \bibinfo{booktitle}{\emph{2019 14th ACM/IEEE International
  Conference on Human-Robot Interaction (HRI)}}. \bibinfo{pages}{191--201}.
\newblock
\urldef\tempurl%
\url{https://doi.org/10.1109/HRI.2019.8673275}
\showDOI{\tempurl}


\end{thebibliography}

\
\end{document}